\documentclass[letterpaper]{article} 
\usepackage{aaai2026}  
\usepackage{times}  
\usepackage{helvet}  
\usepackage{courier}  
\usepackage[hyphens]{url}  
\usepackage{graphicx} 
\usepackage{natbib}  
\usepackage{caption} 
\usepackage{algorithm}
\usepackage{algorithmic}

\usepackage{newfloat}
\usepackage{listings}
\DeclareCaptionStyle{ruled}{labelfont=normalfont,labelsep=colon,strut=off} 
\floatstyle{ruled}
\newfloat{listing}{tb}{lst}{}
\floatname{listing}{Listing}
\usepackage{subcaption}

\usepackage{amsmath}
\usepackage{amssymb}
\usepackage{booktabs}
\usepackage[capitalize]{cleveref}
\usepackage{multirow}

\usepackage{verbatim}

\title{Vision Transformers are Circulant Attention Learners}

\author{
    Dongchen Han\textsuperscript{\rm 1}\equalcontrib,
    Tianyu Li\textsuperscript{\rm 2}\equalcontrib,
    Ziyi Wang\textsuperscript{\rm 1},
    Gao Huang\textsuperscript{\rm 1}\thanks{Corresponding Author.}
}
\affiliations{
    \textsuperscript{\rm 1}Department of Automation, BNRist, Tsinghua University\\
    \textsuperscript{\rm 2}Institute for Interdisciplinary Information Sciences, Tsinghua University\\
}

\usepackage{bibentry}

\begin{document}

\maketitle

\begin{abstract}

The self-attention mechanism has been a key factor in the advancement of vision Transformers. However, its quadratic complexity imposes a heavy computational burden in high-resolution scenarios, restricting the practical application. Previous methods attempt to mitigate this issue by introducing handcrafted patterns such as locality or sparsity, which inevitably compromise model capacity. In this paper, we present a novel attention paradigm termed \textbf{Circulant Attention} by exploiting the inherent efficient pattern of self-attention. Specifically, we first identify that the self-attention matrix in vision Transformers often approximates the Block Circulant matrix with Circulant Blocks (BCCB), a kind of structured matrix whose multiplication with other matrices can be performed in $\mathcal{O}(N\log N)$ time. Leveraging this interesting pattern, we explicitly model the attention map as its nearest BCCB matrix and propose an efficient computation algorithm for fast calculation. The resulting approach closely mirrors vanilla self-attention, differing only in its use of BCCB matrices. Since our design is inspired by the inherent efficient paradigm, it not only delivers $\mathcal{O}(N\log N)$ computation complexity, but also largely maintains the capacity of standard self-attention. Extensive experiments on diverse visual tasks demonstrate the effectiveness of our approach, establishing circulant attention as a promising alternative to self-attention for vision Transformer architectures.

\end{abstract}

\noindent
\textbf{Appendix}: github.com/LeapLabTHU/Circulant-Attention
    
\section{Introduction}

Transformer models have rapidly gained prominence in the field of computer vision in recent years. The superior capacity of self-attention enables vision Transformers to effectively learn from large-scale data, achieving significant success in image classification~\cite{vit}, object detection~\cite{detr}, semantic segmentation~\cite{segformer}, and multimodal tasks~\cite{gsva}. 

However, integrating self-attention into vision architectures also poses a challenge. The quadratic complexity $\mathcal{O}(N^2)$ of self-attention leads to prohibitively high computational cost when applied over a global receptive field. Previous works~\cite{pvt, swin, biformer, flatten, wang2025emulating} address this challenge by introducing \textit{handcrafted} patterns, such as restricting receptive fields or introducing sparsity. While effectively reducing computational demands, these handcrafted designs practically function as \textit{external constraints} imposed on self-attention mechanism, which inevitably compromise long-range modeling capability and limit scalability.

\begin{figure}[t]
    \centering
    \includegraphics[width=0.87\linewidth]{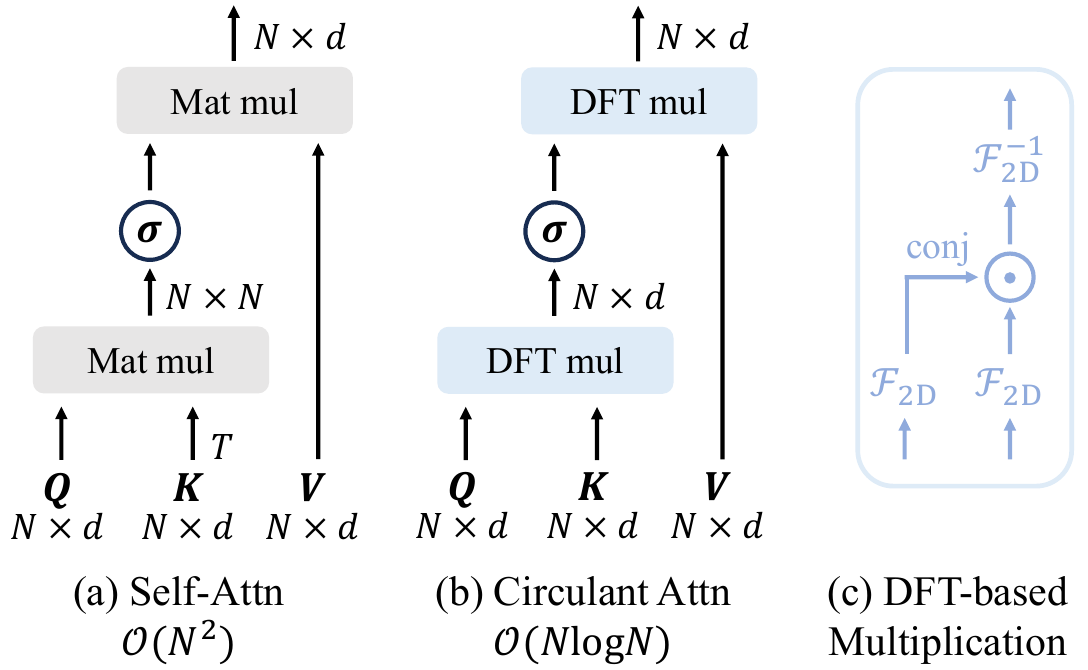}
    \caption{An illustration of vanilla self-attention and the proposed circulant attention. The $\sigma$ represents Softmax function, $\odot$ is the Hadamard product, and $\mathcal{F}_{\rm 2D}, \mathcal{F}_{\rm 2D}^{-1}$ denote the 2D discrete Fourier transform (DFT) and its inversion, respectively. Our circulant attention largely inherits the paradigm of self-attention, except for employing BCCB attention matrices. This simple modification enables our design to be efficiently calculated through DFT-based multiplication, thereby achieving $\mathcal{O}(N\log N)$ complexity. The scaling factor and head-wise summation are omitted for simplicity. Please refer to \cref{sec:method} for details.}
    \label{fig:circulant_attention}
\end{figure}

In this paper, we identify an interesting phenomenon that the attention maps in vision Transformers frequently approximate a kind of special matrix: the Block Circulant matrix with Circulant Blocks (BCCB). This kind of matrix is an extension of the circulant matrix in 2D scenarios, whose multiplication with other matrices can be efficiently implemented by 2D discrete Fourier transform (DFT). This indicates that while standard self-attention operates at $\mathcal{O}(N^2)$ computational cost, it \textit{inherently} learns \textit{efficient} patterns that can be calculated with $\mathcal{O}(N\log N)$ complexity. This motivates us to rethink the design of self-attention and come up with a compelling research question: 

\textit{Can we explicitly set the attention map as a BCCB matrix to facilitate efficient computation, while preserving the high expressiveness of vanilla self-attention?}

To answer this question, we delve into the essence of self-attention operation, presenting a novel paradigm named \textbf{Circulant Attention} to fully exploit the observed pattern. 
Specifically, we explicitly transform attention maps into BCCB matrices by vertically projecting the original self-attention matrices onto the BCCB matrix subspace. 
We demonstrate that this mathematical reformulation enables efficient computation of attention scores and output features via 2D discrete Fourier transform, thus reducing the computation complexity from $\mathcal{O}(N^2)$ to $\mathcal{O}(N\log N)$ with the fast Fourier transform algorithm (FFT).
As illustrated in \cref{fig:circulant_attention}, the resulting method highly resembles vanilla self-attention, except that it directly produces BCCB attention matrices and replaces dense matrix multiplication with DFT-based operations. 
Building on this design, we incorporate a token reweighting module to overcome the inherent limitations of the BCCB structure, which further increases model capacity.

Extensive experiments on diverse visual recognition tasks are conducted to validate the effectiveness of our design, including image classification, object detection, and semantic segmentation. The results confirm that the proposed Circulant Attention offers high efficiency and expressiveness, suggesting it as a promising alternative to self-attention for vision Transformer designs.

Our main contributions and takeaways are as follows:

\begin{itemize}

    \item We reveal that the attention maps in vision Transformers frequently approximate the Block Circulant matrix with Circulant Blocks (BCCB), an extension of the circulant matrix in 2D scenarios. We provide detailed analyses and visualizations of this phenomenon.

    \item We present a novel mechanism dubbed Circulant Attention, which utilizes the observed BCCB pattern to achieve efficient computation with $\mathcal{O}(N\log N)$ complexity. Our method serves as a plug-in attention module and can be applied to various vision Transformer models.

    \item Extensive experiments across image classification, object detection, and semantic segmentation confirm that circulant attention delivers a favorable balance between efficiency and expressiveness, establishing it as a promising alternative to the widely employed self-attention.
\end{itemize}

\section{Related Work}

\textbf{Vision Transformer.} Transformer architectures and attention mechanisms have seen significant advances in computer vision in recent years~\cite{attention}. However, the quadratic computation complexity of self-attention leads to unmanageable cost when processing global feature maps. To address this problem, various approaches have been proposed to reduce the computational overhead by introducing handcrafted patterns, such as locality or sparsity. Local methods such as Swin Transformer~\cite{swin} restrict attention to non-overlapping windows, reducing computational cost. CSwin~\cite{cswin} extends this idea with cross-shaped windows for richer context. Neighborhood Attention Transformer (NAT)~\cite{nat} further mimics convolution by limiting attention to local neighborhoods of each query. Apart from these designs, another line of research employs sparse attention paradigms. PVT~\cite{pvt} employs downsampling of keys and values to reduce computational complexity, and DAT~\cite{dat} presents an input-dependent sparse attention pattern. BiFormer~\cite{biformer} uses bi-level routing attention to dynamically determine areas of interest for each query. Despite their efficiency, these handcrafted attention patterns inevitably compromise the expressiveness of global self-attention. In this paper, we leverage an intrinsic efficient structure of self‑attention to strike a favorable balance between efficiency and expressiveness.

\noindent
\textbf{Efficient architecture with DFT.} The discrete Fourier transform (DFT) has long been an important tool in digital image processing~\cite{digital_image}. Recent work applies DFT and the Fast Fourier Transform (FFT) to design efficient network components~\cite{fno, gfn, afno, aff, freq_deblur}. GFNet~\cite{gfn} utilizes the convolution theorem of DFT to build global depth-wise convolution module, achieving $\mathcal{O}(N\log N)$ complexity via FFT. FNO~\cite{fno} further generalizes this approach to dense global convolution. AFNO~\cite{afno} and AFFNet~\cite{aff} achieve equivalent dynamic depth-wise convolution through input-dependent frequency filters. In this paper, we employ 2D DFT and the FFT algorithm to efficiently implement our circulant attention mechanism.

\newcommand{\vect}[1]{\boldsymbol{#1}}

\newcommand{\FFT}{\mathcal{F}}
\newcommand{\IFFT}{\mathcal{F}^{-1}}

\newcommand{\conj}[1]{\overline{#1}}

\begin{table}[t]
    \centering
    \footnotesize
    \setlength{\tabcolsep}{0.5mm}{
    \renewcommand\arraystretch{1.1}
    \begin{tabular}{c|c}
        \toprule
        \textbf{Notations} & \textbf{Descriptions} \\
        
        \midrule
        $\FFT_{\rm 1D}, \IFFT_{\rm 1D}$ & 1D discrete Fourier transform (DFT) and inversion. \\
        $\FFT_{\rm 2D}, \IFFT_{\rm 2D}$ & 2D discrete Fourier transform (DFT) and inversion. \\

        \midrule
        $\conj{(\cdot)}$ & The complex conjugate. \\
        $\sigma$ & The Softmax operation on each row. \\

        \midrule
        $\odot$ & Hadamard product, element-wise product. \\
        $\circledast$ & The DFT-based matrix multiplication we defined. \\

        \midrule
        $\| \cdot \|$ & The Frobenius norm. \\
        $\langle \cdot, \cdot \rangle$ & The Frobenius inner product. \\

        \bottomrule
    \end{tabular}}
    \caption{Important notations used in this paper.}
    \label{tab:notations}
\end{table}

\section{Preliminaries}
\label{sec:preliminaries}

This section revisits the formulation of self-attention, circulant matrix, and BCCB matrix. To facilitate reading, we summarize important notations in \cref{tab:notations}.

\subsection{Attention Formulation}

We first briefly review the calculation of vanilla self-attention~\cite{attention} in vision Transformer models. Consider an image token sequence $x\in\mathbb{R}^{N\times C}$, where $N=H\times W$ and $H,W,C$ are the height, width, and dimension of the feature map, respectively. In each attention head, $x$ is transformed into $ Q,K,V\in\mathbb{R}^{N\times d} $ through projection matrices $ W_{Q/K/V}\in\mathbb{R}^{C \times d} $, where $d$ is the head dimension. 
Based on this, self-attention computes the attention weights and calculates the output as a weighted sum of values using normalized attention score:
\begin{equation}
\label{eq:self-attention}
    A=QK^{\top}/\sqrt{d},\ \ O=\sigma(A)V,
\end{equation}
where $A\in\mathbb{R}^{N \times N}$ is the raw attention matrix and $\sigma$ represents the Softmax function.

\subsection{Circulant Matrix}


An $N \times N$ matrix $C$ is a circulant matrix if and only if each row is a cyclic shift of the previous one. It has the form:
\begin{equation}
    C = 
    \begin{pmatrix}
    c_0 & c_{1} & \cdots & c_{N-1} \\
    c_{N-1} & c_0 & \cdots & c_{N-2} \\
    \vdots & \vdots & \ddots & \vdots \\
    c_{1} & c_{2} & \cdots & c_0
    \end{pmatrix}.
\end{equation}
This type of matrix can be fully determined by its first row $c = [c_0, c_1, \dots, c_{N-1}]$, where $C_{i,j}=c_{j-i\pmod N}$. 

The multiplication of a circulant matrix $C\in \mathbb R^{N\times N}$ and a vector $x\in \mathbb R^N$ can be expressed as:
\begin{equation}
    (Cx)_i=\sum_{j=0}^{N-1} C_{i,j}x_j=\sum_{j=0}^{N-1}c_{(j-i)\bmod N}\cdot x_j.
\end{equation}

Let $k = (j-i) \bmod N$, which implies $j = (i+k) \bmod N$. We can rewrite the expression as:
\begin{equation}
    (C{x})_i = \sum_{k=0}^{N-1} c_k \cdot x_{(i+k) \bmod N}.
\end{equation}
This is precisely the definition of the 1D circular cross-correlation between $c$ and $x$, which is the 1D depth-wise convolution with circular padding in deep learning (where the concept of convolution does not involve flipping the kernel). Therefore, the multiplication $y = Cx$ can be achieved with the \textit{Cross-Correlation Theorem}~\cite{wang2019kernel}, which states that the Fourier transform of a cross-correlation result is equivalent to the element-wise product of the first signal's conjugated Fourier transform and the second signal's Fourier transform. Mathematically:
\begin{equation}
    \FFT_{\rm 1D}(Cx) = \conj{\FFT_{\rm 1D}({c})} \odot \FFT_{\rm 1D}({x}).
\end{equation}
Thus, we have:
\begin{equation}
    Cx = \IFFT_{\rm 1D}\left( \conj{\FFT_{\rm 1D}({c})} \odot \FFT_{\rm 1D}({x}) \right),
\end{equation}
where $\FFT_{\rm 1D},\IFFT_{\rm 1D}$ denotes the 1D discrete Fourier transform (DFT) and its inversion, $\odot$ is the element-wise (Hadamard) product, and $\conj{(\cdot)}$ represents the complex conjugate. Leveraging the fast Fourier transform algorithm, we can compute $y=Cx$ with a time complexity of $O(N \log N)$.

\subsection{BCCB Matrix}
\label{sec:bccb}


Block Circulant matrix with Circulant Blocks (BCCB) is the 2D generalization of circulant matrix~\cite{davis1979circulant}. A BCCB matrix $B \in \mathbb{R}^{N \times N},N=H\times W$ has a block circulant structure with $H \times H$ blocks:
\begin{equation}
    B = 
    \begin{pmatrix}
    C_0 & C_1 & \cdots & C_{H-1} \\
    C_{H-1} & C_0 & \cdots & C_{H-2} \\
    \vdots & \vdots & \ddots & \vdots \\
    C_1 & C_2 & \cdots & C_0
    \end{pmatrix},
\end{equation}
where each block $C_i$ is a circulant matrix of size $W \times W$. Similar to the circulant matrix, the BCCB matrix $B$ is fully determined by its first row $b = [c_0, c_1, \dots, c_{HW-1}]$. Let $\hat{b},\hat{x}\in \mathbb{R}^{H \times W}$ be the 2D reshaped versions of $b,x$, respectively. As proved in the Appendix, the multiplication $y=Bx$ is equivalent to the 2D circular cross-correlation between $\hat{b}$ and $\hat{x}$, i.e., the 2D depth-wise convolution with circular padding in deep learning. Similar to the 1D scenario, this operation can be implemented by 2D DFT~\cite{davis1979circulant}:
\begin{equation}
    Bx = \IFFT_{\rm 2D}\left( \conj{\FFT_{\rm 2D}(b)} \odot \FFT_{\rm 2D}({x}) \right)\triangleq b\circledast x,
\end{equation}
where $\FFT_{\rm 2D},\IFFT_{\rm 2D}$ denotes the 2D DFT and its inversion. The $\circledast$ is our defined DFT-based multiplication. For simplicity, the reshaping operations between 1D $\mathbb{R}^N$ sequences and 2D $\mathbb{R}^{H \times W}$ feature maps are not demonstrated, and we define the inputs/outputs of $\FFT_{\rm 2D},\IFFT_{\rm 2D}$ to be 1D sequences.

\section{Method}
\label{sec:method}

\begin{figure}[t]
    \centering
    \includegraphics[width=0.95\linewidth]{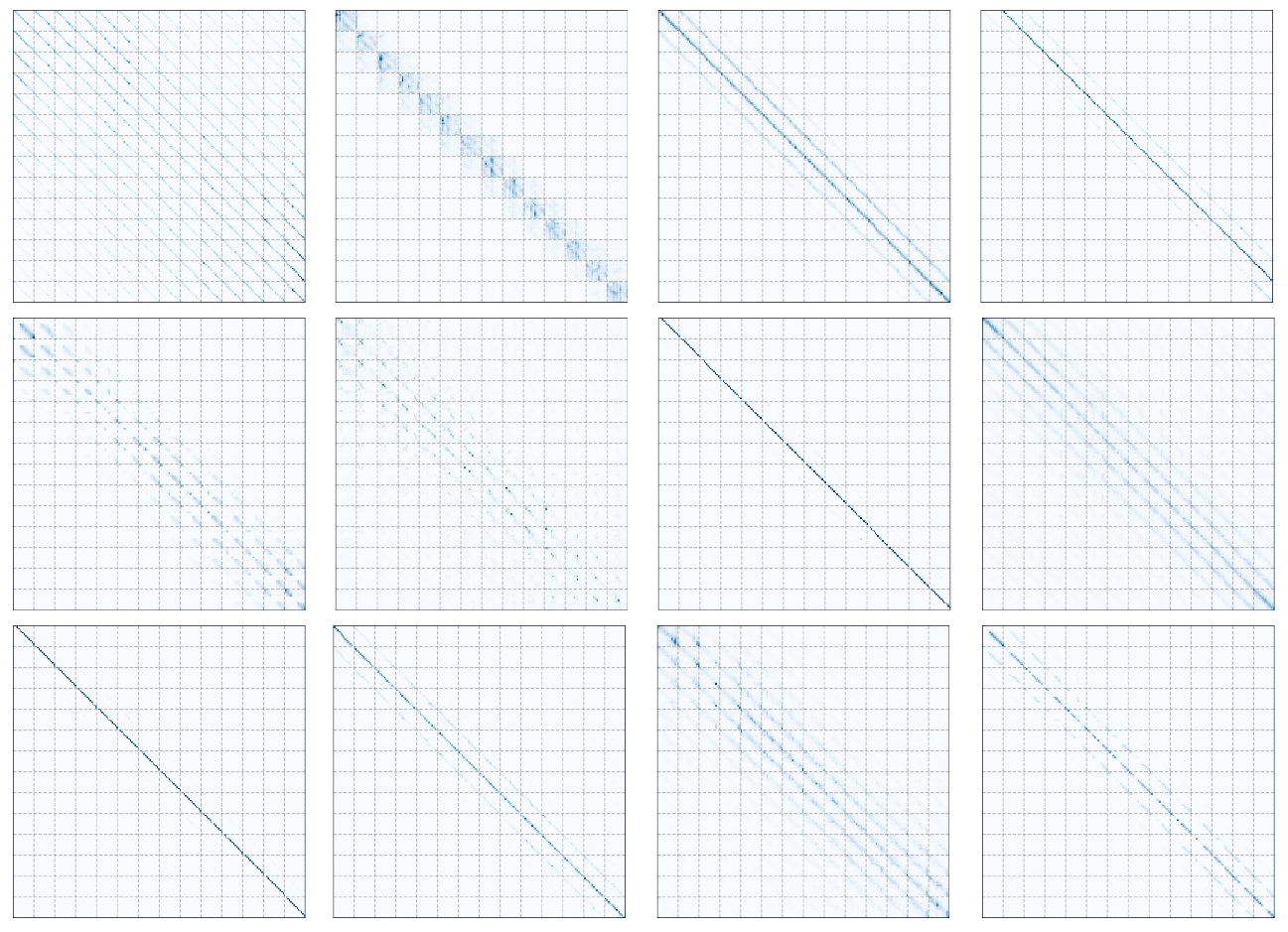}
    \caption{$N\!\times\!N$ attention maps from DeiT~\cite{deit}, where $N=H\!\times\!W, H=W=14$. Without handcrafted constraints, the self-attention module in vision Transformer learns near BCCB patterns. For better observation, we divide the matrix into $H\!\times\!H$ blocks of shape $W\!\times\!W$ using gray dashed lines. Zoom in for best view. }
    \label{fig:BCCB_pattern}
\end{figure}

\subsection{Efficient Pattern in Vision Transformer}

The self-attention mechanism calculates pairwise similarities between each query and key at quadratic complexity $\mathcal{O}(N^2)$. 
However, we find an interesting phenomenon that, in practice, vision Transformer tends to learn attention patterns that are much more structured and efficient. Specifically, in \cref{fig:BCCB_pattern}, we visualize several attention matrices extracted from the DeiT~\cite{deit} model. These matrices closely resemble block circulant matrices with circulant blocks (BCCB). As discussed in \cref{sec:preliminaries}, multiplication by a BCCB matrix is equivalent to a 2D global convolution operation, which can be carried out in $\mathcal{O}(N\log N)$ time via fast Fourier transform algorithm. This suggests that although self-attention formally incurs quadratic complexity, it implicitly inherits an efficient structure.

\begin{figure}[t]
    \centering
    \includegraphics[width=\linewidth]{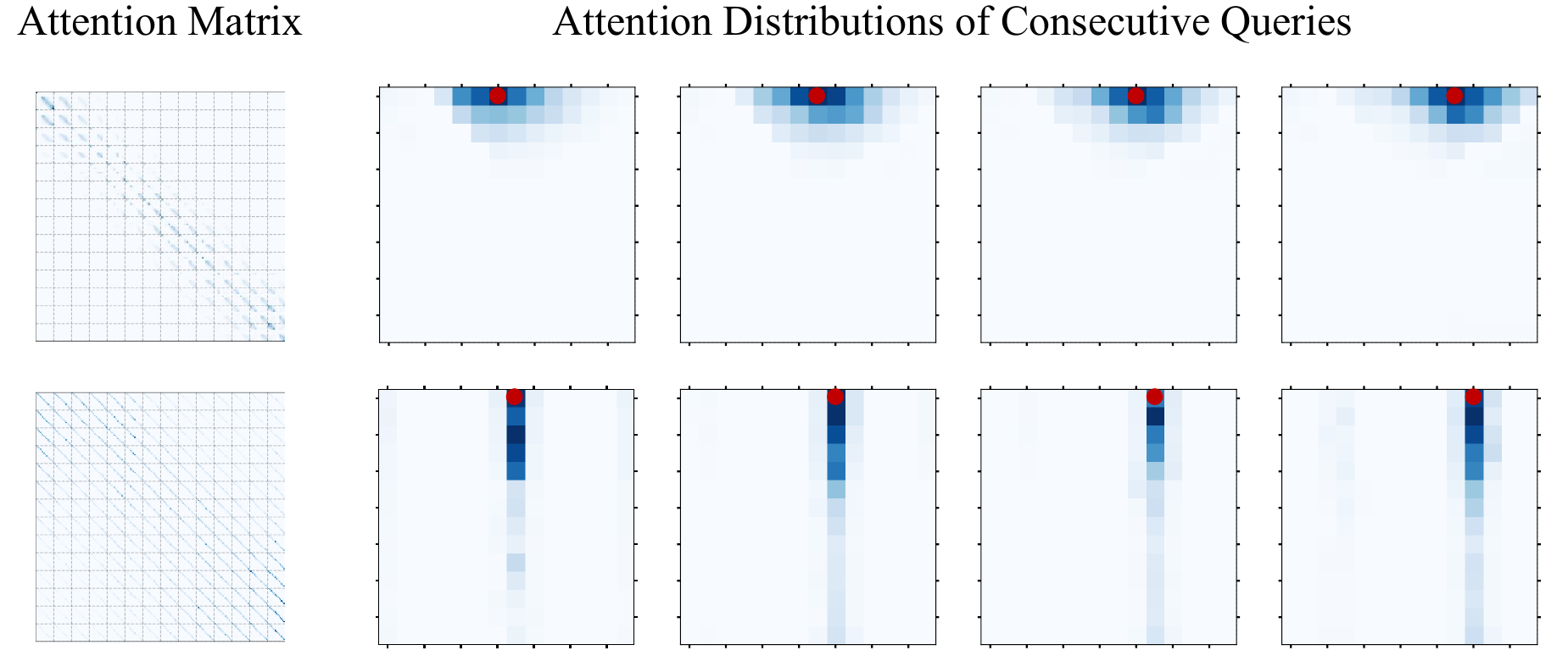}
    \caption{The $N\!\times\!N$ attention matrices and $H\!\times\!W$ attention distributions of consecutive queries, where $N=H\!\times\!W, H=W=14$. The red points correspond to the query tokens. When exhibiting BCCB pattern, the attention distributions show convolution-like translation invariance.}
    \label{fig:BCCB_explain}
\end{figure}

For a better understanding of this phenomenon, we provide the attention distributions of adjacent query tokens (marked in red) in the attention matrices of \cref{fig:BCCB_explain}. 
A key observation is that the attention distributions exhibit an approximate shift invariance corresponding to the movement of the query positions, which is exactly the behavior of a 2D global convolution implemented by a BCCB matrix. This further confirms that the learned attention in vision Transformer exhibits strong similarity to BCCB matrix.

Motivated by the above discoveries, a critical research question emerges: \textit{Can we explicitly enforce a BCCB structure on the attention matrix?}

In the next section, we answer this question with \textbf{Circulant Attention}, a novel attention mechanism that employs BCCB attention matrices. By imposing this structural prior, we aim to explicitly encode the beneficial properties of BCCB matrices, i.e., $\mathcal{O}(N\log N)$ complexity and inherent shift invariance, into the attention mechanism, thereby benefiting from both efficiency and expressiveness.

\subsection{Circulant Attention}


Our circulant attention approximates the original attention map $A\in\mathbb{R}^{N\times N}$ using its nearest Block Circulant with Circulant Blocks (BCCB) matrix $\tilde{A}$. Formally, we have
\begin{equation}
    \tilde{A} = \arg\min_{B \in \mathcal{B}} \|A - B\|,
\end{equation}
where $\mathcal{B}$ denotes the $N\times N$ BCCB matrix subspace and $\| \cdot \|$ represents the Frobenius norm. Therefore, $\tilde{A}$ is the orthogonal projection of $A$ in the BCCB matrix subspace. As discussed in \cref{sec:preliminaries}, a $N\times N$ BCCB matrix could be fully determined by its first row. Let $B_k$ denotes the BCCB matrix whose first row is a one-hot vector with 1 at the $k$-th position. It is easy to see that $\{B_0, \cdots, B_{N-1}\}$ forms the basis for BCCB matrix subspace. Furthermore, we have
\begin{equation}
\label{eq:orthogonal}
    \langle B_k, B_k \rangle=N,\langle B_k, B_j \rangle=0, k\neq j,
\end{equation}
where $\langle \cdot, \cdot \rangle$ denotes the Frobenius inner product, i.e., the sum of element-wise products of two matrices. \cref{eq:orthogonal} indicates that $\{B_0, \cdots, B_{N-1}\}$ is an orthogonal basis. Therefore, the orthogonal projection $\tilde{A}$ in the BCCB matrix subspace can be expressed as:
\begin{equation}
    \begin{split}
        \tilde{A}&=\sum_{k=0}^{N-1} \ \frac{\langle A, B_k \rangle}{\langle B_k, B_k \rangle}\   B_k \\
        &= \frac{1}{N} \sum_{k=0}^{N-1} \ \langle A, B_k \rangle\  B_k \\
        &= \frac{1}{N} \sum_{k=0}^{N-1} \ \langle \frac{QK^\top}{\sqrt{d}}, B_k \rangle\  B_k. \\
    \end{split}
\end{equation}

With the BCCB attention matrix $\tilde{A}$, our circulant attention computes the output as $O=\sigma(\tilde{A})V$, akin to the vanilla self-attention mechanism (see \cref{eq:self-attention}).

\noindent
\textbf{Efficient computation algorithm.} The proposed circulant attention still incurs $\mathcal{O}(N^2)$ complexity when computed using the projection formula above. Here, we demonstrate that our method can be equivalently calculated in $\mathcal{O}(N\log N)$ time employing the 2D DFT. 

Specifically, since $\tilde{A}$ is a BCCB matrix, its entire structure is determined by the first row. We denote this row as $a$, which can be expressed as:
\begin{equation}
    a = \frac{1}{N\sqrt{d}}[\langle QK^\top\!\!, B_0 \rangle, \cdots, \langle QK^\top\!\!, B_{N-1} \rangle].
\end{equation}
According to the properties of the BCCB matrix, $B_k$ actually corresponds to a spatial shift in the 2D space. Define
\begin{equation}
    \Delta h=\lfloor k/W \rfloor,\ \Delta w=k \!\!\!\!\mod W.
\end{equation}
Then we have
\begin{equation} \label{eq:show_conv}
    \begin{split}
        a_k 
        &= \frac{1}{N\sqrt{d}}\langle QK^\top\!\!, B_k \rangle \\
        &= \frac{1}{N\sqrt{d}} \sum_{h=0}^{H-1} \sum_{w=0}^{W-1} \langle \hat{Q}_{h, w}, \hat{K}_{h+\Delta h, w+\Delta w} \rangle,
    \end{split}
\end{equation}
where $\hat{Q},\hat{K}\in\mathbb{R}^{H\times W\times d}$ are the 2D reshaped versions of $Q, K\in\mathbb{R}^{N\times d}$ with circular padding (to handle boundary conditions). It could be observed that $a_k$ is the circular cross-correlation result between $\hat{Q}$ and $\hat{K}$, which is exactly the convolution concept in deep learning (see \cref{sec:preliminaries}). Therefore, $a$ can be computed efficiently using 2D DFT:
\begin{equation}\label{eq:fft_step1}
    \begin{split}
        a&=\frac{1}{N\sqrt{d}}\left[\mathcal{F}_{\rm 2D}^{-1}\left(\overline{\mathcal{F}_{\rm 2D}(Q)}\odot\mathcal{F}_{\rm 2D}(K)\right)\right] \cdot \mathbf{1}_{d\times1} \\
        &=\frac{1}{N\sqrt{d}}(Q\circledast K) \cdot \mathbf{1}_{d\times1},
    \end{split}
\end{equation}
where $\circledast$ is the DFT-based matrix multiplication we defined in \cref{sec:bccb} and $Q\circledast K\in\mathbb{R}^{N\times d}$. The $\mathbf{1}_{d\times1}\in\mathbb{R}^{d\times 1}$ is an all-one vector, which corresponds to the implicit summation over dimension $d$ of the inner product between $\hat{Q}_{h, w}$ and $\hat{K}_{h+\Delta h, w+\Delta w}$ in \cref{eq:show_conv}.

Since $\tilde{A}$ is a BCCB matrix and $\sigma$ is the Softmax function on each row, $\sigma(\tilde{A})$ is also a BCCB matrix and $\sigma(a)$ represents its first row. As discussed in \cref{sec:preliminaries}, the output $O=\sigma(\tilde{A})V$ can thus be efficiently computed with:
\begin{equation}\label{eq:fft_step2}
    \begin{split}
        O&=\mathcal{F}_{\rm 2D}^{-1}\left(\overline{\mathcal{F}_{\rm 2D}(\sigma(a))}\odot\mathcal{F}_{\rm 2D}(V)\right) \\
        &=\sigma(a)\circledast V.
    \end{split}
\end{equation}

\begin{table*}[t]
    \centering
    \footnotesize
    \begin{subtable}[t]{0.475\linewidth}
        \centering
        \setlength{\tabcolsep}{0.5mm}{
        \renewcommand\arraystretch{1.0}
        \begin{tabular}{l|c c c|l}
            \toprule
            \textbf{Method} 
            & \textbf{Reso}    & \textbf{\#Params} & \textbf{FLOPs}    & \textbf{Top-1}\\
            
            \midrule
            DeiT-T~\cite{deit}  
            & ${224}^2$      & 5.7M      & 1.2G      & 72.2\\
            \textbf{CA-DeiT-T} 
            & ${224}^2$      &    6.1M       &    1.2G       & \textbf{75.0\,{\scriptsize (+2.8)}}\\
            DeiT-S~\cite{deit}
            & ${224}^2$      & 22.1M     & 4.6G      & 79.8\\
            \textbf{CA-DeiT-S} 
            & ${224}^2$      &     23.8M      &   4.8G        & \textbf{81.0\,{\scriptsize (+1.2)}}\\
            DeiT-B~\cite{deit}
            & ${224}^2$      & 86.6M     & 17.6G     & 81.8\\
            \textbf{CA-DeiT-B}
            & ${224}^2$      &     93.6M      &  18.9G         & \textbf{82.3\,{\scriptsize (+0.5)}}\\
            
            \midrule
            PVT-T~\cite{pvt}  
            & ${224}^2$      & 13.2M     & 1.9G      & 75.1\\
            \textbf{CA-PVT-T} 
            & ${224}^2$      &      12.2M     &     2.0G      & \textbf{78.1\,{\scriptsize (+3.0)}}\\
            PVT-S~\cite{pvt}
            & ${224}^2$      & 24.5M     & 3.8G      & 79.8\\
            \textbf{CA-PVT-S} 
            & ${224}^2$      &    22.8M       &     4.0G      & \textbf{81.7\,{\scriptsize (+1.9)}}\\
            PVT-M~\cite{pvt}
            & ${224}^2$      & 44.2M     & 6.7G      & 81.2\\
            \textbf{CA-PVT-M} 
            & ${224}^2$      &     42.5M      &     6.8G      & \textbf{82.6\,{\scriptsize (+1.4)}}\\
            PVT-L~\cite{pvt}
            & ${224}^2$      & 61.4M     & 9.8G      & 81.7\\
            \textbf{CA-PVT-L} 
            & ${224}^2$      &   58.6M       &       10.1G    & \textbf{82.9\,{\scriptsize (+1.2)}}\\
            
            \midrule
            Swin-T~\cite{swin}  
            & ${224}^2$      & 29M       & 4.5G      & 81.3\\
             \textbf{CA-Swin-T} 
            & ${224}^2$      & 28M       & 4.6G      & \textbf{82.2\,{\scriptsize (+0.9)}}\\
            Swin-S~\cite{swin}
            & ${224}^2$      & 50M       & 8.7G      & 83.0\\
             \textbf{CA-Swin-S} 
            & ${224}^2$      & 50M       & 8.8G      & \textbf{83.6\,{\scriptsize (+0.6)}}\\
            Swin-B~\cite{swin}
            & ${224}^2$      & 88M       & 15.4G     & 83.5\\
           \textbf{CA-Swin-B} 
            & ${224}^2$      & 88M       & 15.7G     & \textbf{83.9\,{\scriptsize (+0.4)}}\\
            Swin-B~\cite{swin}
            & ${384}^2$      & 88M       & 47.0G     & 84.5\\
             \textbf{CA-Swin-B} 
            & ${384}^2$      & 88M       & 47.1G     & \textbf{85.1\,{\scriptsize (+0.6)}}\\
            \bottomrule
        \end{tabular}}
    \end{subtable}
    \begin{subtable}[t]{0.515\linewidth}
        \centering
        \setlength{\tabcolsep}{0.5mm}{
        \renewcommand\arraystretch{1.0}
        \begin{tabular}{l|c c c|l}
            \toprule
            \textbf{Method} 
            & \textbf{Reso}    & \textbf{\#Params} & \textbf{FLOPs}    & \textbf{Top-1}\\
            
            \midrule
            SLAB-T~\cite{slab}
            & ${224}^2$     & 29M       & 4.5G      & 81.8 \\
            VMamba-T~\cite{vmamba} 
            & ${224}^2$     & 31M       & 4.9G      & 82.5 \\
            SOFT-S++~\cite{soft_plus}
            & ${224}^2$     & 27M       & 4.5G      & 82.6 \\
            PolaFormer-T~\cite{polaformer}
            & ${224}^2$     & 29M       & 4.5G      & 82.6 \\
            LocalVMamba-T~\cite{localmamba} 
            & ${224}^2$     & 26M       & 5.7G      & 82.7 \\
            VVT-S~\cite{vvt}
            & ${224}^2$     & 26M       & 5.6G      & 82.7 \\
            Agent-CSwin-T~\cite{agent_attention}
            & ${224}^2$     & 21M       & 4.3G      & 83.1 \\
            FasterViT-1~\cite{fastervit}
            & ${224}^2$     & 53M       & 5.3G      & 83.2 \\
            EfficientViT-B3~\cite{efficientvit}
            & ${224}^2$     & 49M       & 4.0G      & 83.5 \\
             \textbf{CAT-T}
            & ${224}^2$     & 27M       & 4.3G      & \textbf{83.6}\\
            
            \midrule
            LocalVMamba-S~\cite{localmamba} 
            & ${224}^2$     & 50M       & 11.4G     & 83.7 \\
            QFormer$_h$-S~\cite{quadrangle} 
            & ${224}^2$     & 51M       & 8.9G      & 84.0 \\
            BiFormer-B~\cite{biformer}
            & $224^2$       & 57M       & 9.8G      & 84.3 \\
            MILA-S~\cite{demystify_mamba}
            & ${224}^2$     & 43M       & 7.3G      & 84.4 \\
            TransXNet-B~\cite{transxnet}
            & ${224}^2$     & 48M       & 8.3G      & 84.6 \\
            \textbf{CAT-S} 
            & ${224}^2$     & 51M       & 7.9G      & \textbf{84.5}\\
            
            \midrule
            VMamba-B~\cite{vmamba} 
            & ${224}^2$     & 89M       & 15.4G     & 83.9 \\
            InLine-Swin-B~\cite{inline}
            & ${224}^2$     & 88M       & 15.4G     & 84.1 \\
            SOFT-L++~\cite{soft_plus}
            & ${224}^2$     & 85M       & 15.4G     & 84.1 \\
            FasterViT-3~\cite{fastervit}
            & ${224}^2$     & 160M      & 18.2G     & 84.9 \\
            OverLock-B~\cite{overlock}
            & ${224}^2$     & 95M       & 16.7G     & 85.1 \\
           \textbf{CAT-B} 
            & ${224}^2$     & 90M       & 15.2G     & \textbf{85.0}\\
            \bottomrule
        \end{tabular}}
    \end{subtable}
    \caption{Comparison with baseline models (left) and advanced methods (right) on ImageNet-1K.}
    \label{tab:imagenet}
\end{table*}

In \cref{fig:circulant_attention}, we provide an illustration of the proposed circulant attention paradigm. It can be observed that our design is structurally similar to vanilla self-attention, with the key differences being the use of BCCB attention matrices and DFT-based efficient multiplication.

\noindent
\textbf{Complexity analysis.}
Leveraging the efficient computation algorithm, the overall complexity of our circulant attention is expressed as:
\begin{equation}
    \begin{split}
        \Omega(\mathrm{CA})=&\underbrace{2N(\log_2\!N)d+2Nd+N\log_2\!N}_{\text{DFT, Product, IDFT in Eq. (7)}}+ \\
        &\underbrace{N(\log_2\!N)(d+1)+2Nd+N(\log_2\!N)d}_{\text{DFT, Product, IDFT in Eq. (8)}} \\
        =&N(\log_2\!N)(4d+2)+4Nd,
    \end{split}
\end{equation}
which is much more efficient than the computation complexity of vanilla self-attention:
\begin{equation}
    \Omega(\mathrm{SA})=N^2d+N^2d=2N^2d.
\end{equation}

\noindent
\textbf{Token reweighting module.} 
In standard self-attention, a row-wise Softmax is applied to the raw attention map $A\in\mathbb{R}^{N \times N}$, ensuring that each row sums to one. Notably, this operation does not constrain the column sums of $\sigma(A)$, allowing certain keys to accumulate higher total attention scores across queries. Consequently, different queries can emphasize similar keys and values, helping the model distinguish more informative tokens from others.
However, the block circulant structure of our circulant attention map $\sigma(\tilde{A})$ enforces both row and column sums to equal one, thereby limiting its ability to highlight salient tokens.
As a remedy, we introduce a simple yet effective token reweighting method to further improve our design. Specifically, there are two ways to implement this design: the pre-reweighting
\begin{equation} \label{eq:pre_reweighting}
    O=\mathrm{CirAttn}(Q, K, V\odot T),
\end{equation}
and the post-reweighting
\begin{equation} \label{eq:post_reweighting}
    O=\mathrm{CirAttn}(Q, K, V)\odot T,
\end{equation}
where $T=\mathrm{SiLU}(xW_T)\in\mathbb{R}^{N \times d}$ is an input-dependent token reweighting factor, and $\mathrm{CirAttn}$ represents our circulant attention operator.

\subsection{Implementation}

Our circulant attention serves as a plug-in module and can be applied to various vision Transformer models. As a showcase, we employ three representative Transformer architectures: DeiT~\cite{deit}, PVT~\cite{pvt}, and Swin Transformer~\cite{swin} to implement our approach. These three models represent global self-attention, sparse attention and local attention, respectively. We replace the original attention module in these models with circulant attention to establish our models. 
Since our method benefits from $N\log N$ complexity, it is possible to directly process a high-resolution feature map in the early stages without incurring high computational cost. Therefore, for hierarchical models like Swin and PVT, we mainly restrict the attention replacement to the first two stages.
Beyond baseline improvements, we have also developed a family of specialized models, termed Circulant Attention Transformer (CAT), to compare with various advanced vision Transformers. Detailed model architectures are shown in the Appendix.

\begin{table}[t]
\centering
\footnotesize
\setlength{\tabcolsep}{0.9mm}{
\renewcommand\arraystretch{1.0}
\begin{tabular}{l|c|c|ccc|ccc}
    \toprule
    \multicolumn{9}{c}{\textbf{Mask R-CNN Object Detection on COCO}} \\
    Backbone & FLOPs & Sch. & AP$^b$ & AP$^b_\text{50}$ & AP$^b_\text{75}$ & AP$^m$ & AP$^m_\text{50}$ & AP$^m_\text{75}$ \\
    
    \midrule[0.8pt]
    PVT-T 
    & 240G & 1x      & 36.7 & 59.2 & 39.3 & 35.1 & 56.7 & 37.3 \\
    \textbf{CA-PVT-T}
    & 218G & 1x      & 40.5 & 63.4 & 43.9 & 37.9 & 60.4 & 40.7 \\

    \hline
    PVT-S     
    & 305G & 1x      & 40.4 & 62.9 & 43.8 & 37.8 & 60.1 & 40.3 \\
    \textbf{CA-PVT-S}
    & 269G & 1x      & 44.2 & 66.9 & 48.3 & 40.7 & 63.9 & 43.7 \\

    \hline
    PVT-M     
    & 392G & 1x      & 42.0 & 64.4 & 45.6 & 39.0 & 61.6 & 42.1 \\
    \textbf{CA-PVT-M}
    & 356G & 1x      & 45.2 & 67.7 & 49.4 & 41.2 & 64.7 & 44.1 \\

    \hline
    PVT-L 
    & 494G & 1x      & 42.9 & 65.0 & 46.6 & 39.5 & 61.9 & 42.5 \\
    \textbf{CA-PVT-L}
    & 444G & 1x      & 46.3 & 68.8 & 50.8 & 41.9 & 65.4& 45.0 \\

    \midrule[0.8pt]
    Swin-T &267G&1x&43.7 &66.6& 47.7& 39.8& 63.3& 42.7\\
    \textbf{CA-Swin-T}& 269G&1x &44.5& 67.3 &48.9& 40.5& 64.2& 43.5\\
    \hline
    Swin-S &358G& 1x& 45.7& 67.9& 50.4& 41.1& 64.9& 44.2\\
    \textbf{CA-Swin-S} & 361G & 1x& 46.8& 69.3& 51.7& 42.2& 66.2& 45.3\\

    \hline
    Swin-B
    & 503G & 1x      & 46.9 & -    & -    & 42.3 & -    & -    \\
    \textbf{CA-Swin-B}&507G & 1x&47.5& 70.1& 52.2&  42.8& 66.8 &45.9\\
    \hline
    Swin-T& 267G& 3x &46.0& 68.1& 50.3& 41.6& 65.1 &44.9\\
    \textbf{CA-Swin-T}&269G & 3x&46.7& 68.8& 51.2&  42.5& 65.9 &45.9\\
    \toprule
\end{tabular}}
\caption{Results on COCO dataset. The FLOPs are computed with an input resolution of 1280$\times$800.}
\label{tab:COCO}
\end{table}

\begin{table}[t]
\centering
\footnotesize
\setlength{\tabcolsep}{2.5mm}{
\renewcommand\arraystretch{1.0}
\begin{tabular}{l|c|c c|c}
    \toprule
    \multicolumn{5}{c}{\textbf{Semantic Segmentation on ADE20K}} \\
    Backbone & Method & FLOPs & \#Params & mIoU \\
    
    \midrule[0.8pt]
    PVT-T 
    & S-FPN & 158G & 17M & 35.7 \\
    \textbf{CA-PVT-T}
    & S-FPN &135G & 16M& 39.4 \\

    \hline
    PVT-S 
    & S-FPN & 225G & 28M & 39.8 \\
    \textbf{CA-PVT-S}
    & S-FPN & 187G & 26M & 42.3 \\

    \hline
    PVT-M 
    & S-FPN & 315G & 48M & 41.6 \\
    \textbf{CA-PVT-M}
    & S-FPN & 278G & 46M & 43.7 \\

    \hline
    PVT-L 
    & S-FPN & 420G & 65M & 42.1 \\
    \textbf{CA-PVT-L}
    & S-FPN & 369G & 62M & 44.2 \\

    \midrule[0.8pt]
    Swin-T 
    & UperNet & 945G & 60M & 44.5 \\
    \textbf{CA-Swin-T}
    & UperNet & 947G & 59M & 45.2 \\

    \hline
    Swin-S 
    & UperNet & 1038G & 81M & 47.6 \\
    \textbf{CA-Swin-S}
    & UperNet & 1040G & 80M & 48.6 \\
    
    \bottomrule
\end{tabular}}
\caption{Results of semantic segmentation. The FLOPs are computed over encoders and decoders with an input image at the resolution of 512$\times$2048. S-FPN is short for SemanticFPN \cite{semfpn} model.}
\label{tab:ade20k}
\end{table}

\section{Experiments}

In this section, we conduct extensive experiments to evaluate the effectiveness of our method, including ImageNet classification~\cite{imagenet}, COCO object detection~\cite{coco}, and ADE20K semantic segmentation~\cite{ade20k}. Visualization and analysis are also provided.

\subsection{  ImageNet Classification}

The ImageNet-1K~\cite{imagenet} dataset comprises 1.28 million training and 50 thousand validation images spanning 1,000 classes. To ensure a fair comparison, we adopt identical training settings as the baseline models. Specifically, models are trained from scratch for 300 epochs using the AdamW~\cite{adamw} optimizer, with cosine learning rate decay and a linear warm-up over the first 20 epochs. The initial learning rate is set to $1 \times 10^{-3}$, and the weight decay is 0.05. Augmentation and regularization strategies include RandAugment~\cite{randaugment}, Mixup~\cite{mixup}, CutMix~\cite{cutmix} and random erasing~\cite{random_erasing}.

From \cref{tab:imagenet} left, we observe that replacing self-attention in the three representative models with circulant attention leads to consistent improvements. For example, the CA-DeiT outperforms the global self-attention baseline DeiT~\cite{deit}. This indicates that our method not only enjoys high efficiency, but also facilitates the learning of vision Transformers. Furthermore, compared to the sparse attention PVT~\cite{pvt} and local attention Swin Transformer~\cite{swin}, circulant attention demonstrates significant advantages. Our CA-PVT-S matches PVT‑L’s accuracy using 30\% of the parameters and 40\% of the FLOPs. These results establish circulant attention as a promising alternative to self-attention.

\cref{tab:imagenet} right compares our CAT model against various state‑of‑the‑art vision Transformers and Mamba variants. We see that circulant attention delivers better results than highly optimized methods, validating the superior capacity of our approach.

\begin{figure}[t!]
    \centering
    \includegraphics[width=\linewidth]{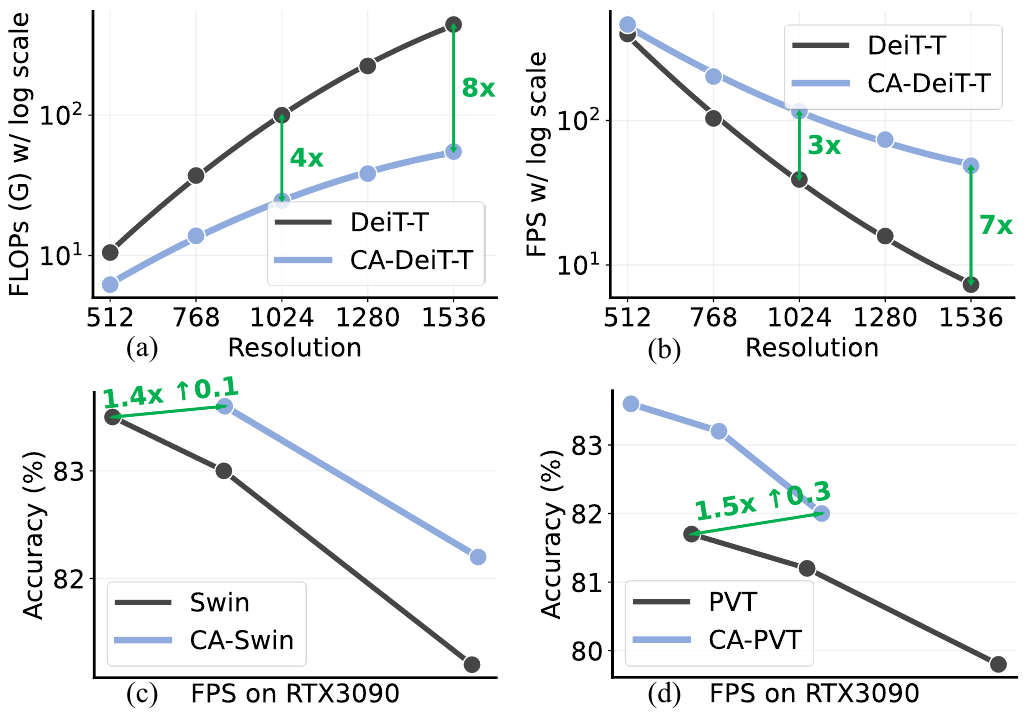}
    \caption{Comparisons between self-attention and the proposed circulant attention in (a) FLOPs, (b) inference FPS, and (c, d) throughput-accuracy trade-off. Throughput is measured on a RTX3090 GPU. }
    \label{fig:speed}
\end{figure}

\begin{figure*}[t!]
    \centering
    \includegraphics[width=0.94\linewidth]{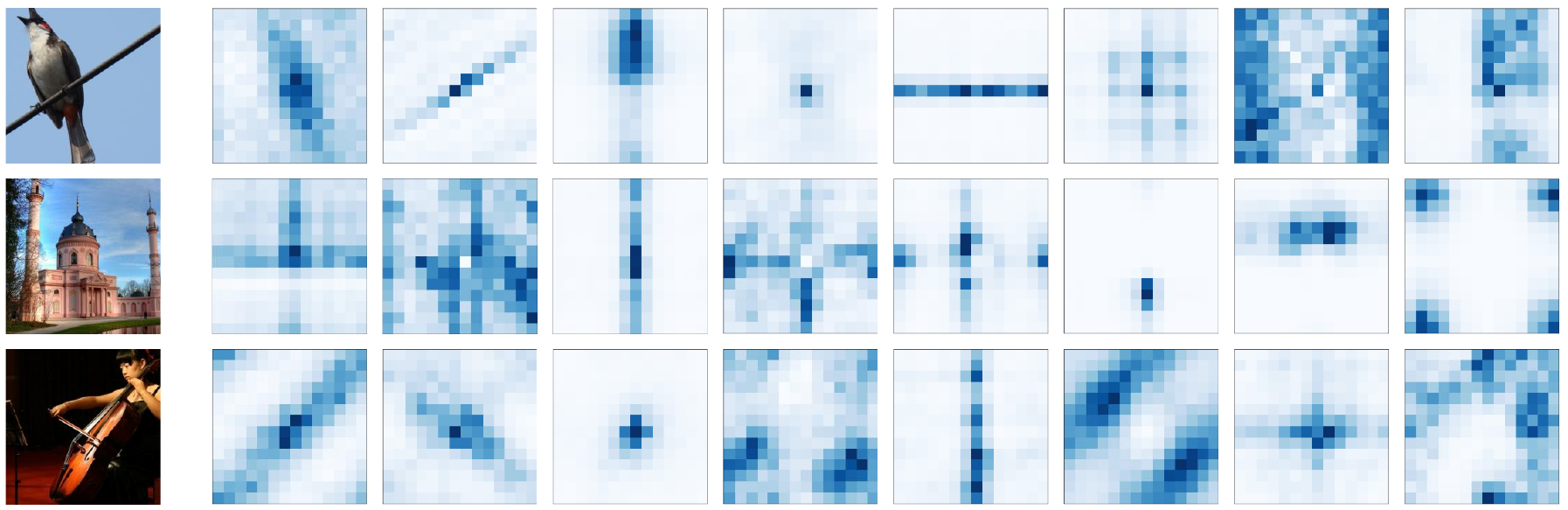}
    \caption{An illustration of the equivalent global convolution kernels from CA-DeiT model.}
    \label{fig:kernels}
\end{figure*}

\subsection{Object Detection}

The COCO~\cite{coco} object detection and instance segmentation dataset contains 118K training images and 5K validation images. We follow the training and testing strategies of the baseline models.
The results are shown in \cref{tab:COCO}. Circulant attention offers effective global modeling with $\mathcal{O}(N\log N)$ complexity, making it ideally suitable for high-resolution image modeling scenarios. Notably, CA-PVT-S outperforms the larger PVT-L model by 1.3 box AP with substantially fewer FLOPs. This gain is more pronounced than in the classification task, where CA-PVT-S and PVT-L yield the same accuracy. These results highlight the superiority of our method in high-resolution scenarios.


\subsection{Semantic Segmentation}
ADE20K~\cite{ade20k} is a well-established benchmark for semantic segmentation, consisting of 20K training images, 2K validation images, and 150 semantic categories. We use the same settings as baseline models.
Similar to the object detection task, the results in Table~\ref{tab:ade20k} demonstrate that circulant attention consistently enhances the performance of PVT and Swin models. 
We observe up to 3.7\% mIoU gain with comparable or less computation cost and parameters.

\subsection{Analysis and Visualization}

\noindent
\textbf{Efficiency analysis.}
Benefiting from $\mathcal{O}(N\log N)$ complexity, our circulant attention shows remarkable efficiency advantages over self-attention operation. 
As demonstrated in \cref{fig:speed}(a), CA-DeiT-T requires $8\times$ fewer FLOPs at a $1536^2$ image resolution, highlighting its potential for high-resolution modeling tasks.
Moreover, these theoretical savings directly translate into practical gains, with our method delivering a $7\times$ speedup at $1536^2$ resolution in \cref{fig:speed}(b). 
Additionally, \cref{fig:speed}(c, d) shows that our model achieves a better trade-off between throughput and accuracy at the default ImageNet resolution $224^2$, enjoying up to $1.5\times$ faster inference speed with improved performance.

\noindent
\textbf{Visualization.}
As discussed in \cref{sec:bccb}, the BCCB matrices correspond to 2D global convolution operations. Therefore, it is easy to visualize and interpret the equivalent global convolution kernels. As shown in \cref{fig:kernels}, our model can generate diverse equivalent kernels based on the input image. For example, given the first image, our model generates bird- and wire-shaped kernels (the first two). There are also kernels focusing more on spatial patterns, such as local, global, half-plane, strip, and cross-shaped ones.

\begin{table}[t]
    \centering
    \footnotesize
    \setlength{\tabcolsep}{1.0mm}{
    \renewcommand\arraystretch{1.1}
    \begin{tabular}{l|c c|c c}
        \toprule
        \                               & FLOPs     & \#Param   & Acc.          & Diff. \\
        \midrule
        DeiT-S~\cite{deit}              & 4.6G      & 22M       & 79.8          & -1.2 \\
        ~~~ $+$ \ \ Circulant Attention   & 4.4G      & 22M       & 79.7          & -1.3 \\
        ~~~ $+$ \ \ Head Dim $d=1$        & 4.4G      & 22M       & 80.2          & -0.8 \\
        %
        \multirow{2}{*}{~~~ $+$ $\begin{array}{l} \text{Token Pre-Reweighting} \\ \text{Token Post-Reweighting} \end{array}$}
        & 4.8G      & 24M       & 80.9          & -0.1 \\
        & 4.8G      & 24M       & \textbf{81.0}          &\textbf{ Ours} \\
        \midrule
        DeiT-S $+$ Token Reweighting        & 5.0G      & 24M       & 80.0          & -1.0 \\
        \bottomrule
    \end{tabular}}
    \caption{Ablation on the key designs based on DeiT-S.}
    \label{tab:ablation}
\end{table}

\subsection{Ablation Study}

We conduct ablation studies to assess the contribution of each component in our circulant attention design. Experiments are performed on ImageNet‑1K under the CA‑DeiT framework.
As shown in \cref{tab:ablation}, we begin with the DeiT-S baseline and introduce our designs in turn. Simply introducing circulant attention leads to a negligible 0.1\% performance drop, indicating that the BCCB pattern is suitable for vision Transformers and does not sacrifice expressiveness. 
On this basis, we find that our design benefits from a small head dimension and more heads. Setting the head dimension $d=1$ improves the accuracy to 80.2\%, outperforming the DeiT-S baseline. This can be attributed to the fact that circulant attention score is a summation of $N$ query-key pairs (see \cref{eq:show_conv}), thus having an equivalent head dimension of $Nd$, which is still large enough when $d=1$.
Additionally, we study the two token reweighting methods defined in \cref{eq:pre_reweighting} and \cref{eq:post_reweighting}. Both designs lead to further improvement, while the post-reweighting delivers a slightly better result. 
Notably, token reweighting offers limited gains on the DeiT-S baseline.
In this paper, we employ post-reweighting as default, achieving 81.0\% accuracy on DeiT-S.
These results confirm the effectiveness of our design.

\section{Conclusion}

This paper reveals an interesting pattern in vision Transformers, where the self-attention maps frequently exhibit nearly block circulant with circulant blocks (BCCB) structures. This observation directly inspires our design of Circulant Attention, a novel attention paradigm that explicitly leverages this inherent pattern to optimize computational efficiency while preserving expressive power. Extensive experiments across classification, object detection, and semantic segmentation fully demonstrate the effectiveness of the proposed circulant attention, establishing it as an efficient and competitive alternative to widely adopted self-attention for vision Transformer architectures.

\section*{Acknowledgements}
This work is supported in part by the National Key R\&D Program of China under Grant 2024YFB4708200, the National Natural Science Foundation of China under Grants U24B20173 and 42327901, and the Scientific Research Innovation Capability Support Project for Young Faculty under Grant ZYGXQNJSKYCXNLZCXM-I20.

\bibliography{main}

@String(IJCV = {Int. J. Comput. Vis.})

@String(CVPR= {IEEE Conf. Comput. Vis. Pattern Recog.})

@String(ICCV= {Int. Conf. Comput. Vis.})

@String(ECCV= {Eur. Conf. Comput. Vis.})

@String(ICLR = {Int. Conf. Learn. Represent.})

@String(AAAI = {AAAI})

@String(CVPRW= {IEEE Conf. Comput. Vis. Pattern Recog. Worksh.})

@String(IJCV  = {IJCV})

@String(CVPR  = {CVPR})

@String(ICCV  = {ICCV})

@String(ECCV  = {ECCV})

@String(ICLR  = {ICLR})

@String(CVPRW= {CVPRW})

@book{davis1979circulant,
  author    = "Davis, Philip J.",
  title     = "Circulant Matrices",
  year      = 1979,
  publisher = "John Wiley \& Sons",
  isbn      = "9780471057710",
}

@inproceedings{deit,
  title={Training data-efficient image transformers \& distillation through attention},
  author={Touvron, Hugo and Cord, Matthieu and Douze, Matthijs and Massa, Francisco and Sablayrolles, Alexandre and J{\'e}gou, Herv{\'e}},
  booktitle={ICML},
  year={2021}
}

@inproceedings{pvt,
  title={Pyramid vision transformer: A versatile backbone for dense prediction without convolutions},
  author={Wang, Wenhai and Xie, Enze and Li, Xiang and Fan, Deng-Ping and Song, Kaitao and Liang, Ding and Lu, Tong and Luo, Ping and Shao, Ling},
  booktitle={ICCV},
  year={2021}
}

@inproceedings{swin,
  title={Swin transformer: Hierarchical vision transformer using shifted windows},
  author={Liu, Ze and Lin, Yutong and Cao, Yue and Hu, Han and Wei, Yixuan and Zhang, Zheng and Lin, Stephen and Guo, Baining},
  booktitle={ICCV},
  year={2021}
}

@inproceedings{cswin,
  title={Cswin transformer: A general vision transformer backbone with cross-shaped windows},
  author={Dong, Xiaoyi and Bao, Jianmin and Chen, Dongdong and Zhang, Weiming and Yu, Nenghai and Yuan, Lu and Chen, Dong and Guo, Baining},
  booktitle={CVPR},
  year={2022}
}

@inproceedings{vit,
  title={An image is worth 16x16 words: Transformers for image recognition at scale},
  author={Dosovitskiy, Alexey and Beyer, Lucas and Kolesnikov, Alexander and Weissenborn, Dirk and Zhai, Xiaohua and Unterthiner, Thomas and Dehghani, Mostafa and Minderer, Matthias and Heigold, Georg and Gelly, Sylvain and others},
  booktitle={ICLR},
  year={2021}
}

@inproceedings{dat,
  title={Vision transformer with deformable attention},
  author={Xia, Zhuofan and Pan, Xuran and Song, Shiji and Li, Li Erran and Huang, Gao},
  booktitle={CVPR},
  year={2022}
}

@inproceedings{detr,
  title={End-to-end object detection with transformers},
  author={Carion, Nicolas and Massa, Francisco and Synnaeve, Gabriel and Usunier, Nicolas and Kirillov, Alexander and Zagoruyko, Sergey},
  booktitle={ECCV},
  year={2020}
}

@inproceedings{segformer,
  title={SegFormer: Simple and efficient design for semantic segmentation with transformers},
  author={Xie, Enze and Wang, Wenhai and Yu, Zhiding and Anandkumar, Anima and Alvarez, Jose M and Luo, Ping},
  booktitle={NeurIPS},
  year={2021}
}

@inproceedings{nat,
  title={Neighborhood attention transformer},
  author={Hassani, Ali and Walton, Steven and Li, Jiachen and Li, Shen and Shi, Humphrey},
  booktitle={CVPR},
  year={2023}
}

@inproceedings{imagenet,
  title={Imagenet: A large-scale hierarchical image database},
  author={Deng, Jia and Dong, Wei and Socher, Richard and Li, Li-Jia and Li, Kai and Fei-Fei, Li},
  booktitle={CVPR},
  year={2009}
}

@inproceedings{adamw,
  title={Decoupled weight decay regularization},
  author={Loshchilov, Ilya and Hutter, Frank},
  booktitle={ICLR},
  year={2018}
}

@inproceedings{randaugment,
  title={Randaugment: Practical automated data augmentation with a reduced search space},
  author={Cubuk, Ekin D and Zoph, Barret and Shlens, Jonathon and Le, Quoc V},
  booktitle={CVPRW},
  year={2020}
}

@inproceedings{cutmix,
  title={Cutmix: Regularization strategy to train strong classifiers with localizable features},
  author={Yun, Sangdoo and Han, Dongyoon and Oh, Seong Joon and Chun, Sanghyuk and Choe, Junsuk and Yoo, Youngjoon},
  booktitle={ICCV},
  year={2019}
}

@inproceedings{mixup,
    title={mixup: Beyond Empirical Risk Minimization},
    author={Hongyi Zhang and Moustapha Cisse and Yann N. Dauphin and David Lopez-Paz},
    booktitle={ICLR},
    year={2018}
}

@inproceedings{random_erasing,
  title={Random erasing data augmentation},
  author={Zhong, Zhun and Zheng, Liang and Kang, Guoliang and Li, Shaozi and Yang, Yi},
  booktitle={AAAI},
  year={2020}
}

@inproceedings{coco,
  title={Microsoft coco: Common objects in context},
  author={Lin, Tsung-Yi and Maire, Michael and Belongie, Serge and Hays, James and Perona, Pietro and Ramanan, Deva and Doll{\'a}r, Piotr and Zitnick, C Lawrence},
  booktitle={ECCV},
  year={2014},
}

@article{ade20k,
  title={Semantic understanding of scenes through the ade20k dataset},
  author={Zhou, Bolei and Zhao, Hang and Puig, Xavier and Xiao, Tete and Fidler, Sanja and Barriuso, Adela and Torralba, Antonio},
  journal={IJCV},
  year={2019}
}

@inproceedings{semfpn,
  title={Panoptic feature pyramid networks},
  author={Kirillov, Alexander and Girshick, Ross and He, Kaiming and Doll{\'a}r, Piotr},
  booktitle={CVPR},
  year={2019}
}

@inproceedings{attention,
  title={Attention is all you need},
  author={Vaswani, Ashish and Shazeer, Noam and Parmar, Niki and Uszkoreit, Jakob and Jones, Llion and Gomez, Aidan N and Kaiser, {\L}ukasz and Polosukhin, Illia},
  booktitle={NeurIPS},
  year={2017}
}

@InProceedings{flatten,
  title={FLatten Transformer: Vision Transformer using Focused Linear Attention},
  author={Han, Dongchen and Pan, Xuran and Han, Yizeng and Song, Shiji and Huang, Gao},
  booktitle={ICCV},
  year={2023}
}

@inproceedings{biformer,
  title={BiFormer: Vision Transformer with Bi-Level Routing Attention},
  author={Zhu, Lei and Wang, Xinjiang and Ke, Zhanghan and Zhang, Wayne and Lau, Rynson WH},
  booktitle={CVPR},
  year={2023}
}

@inproceedings{demystify_mamba,
  title={Demystify Mamba in Vision: A Linear Attention Perspective},
  author={Han, Dongchen and Wang, Ziyi and Xia, Zhuofan and Han, Yizeng and Pu, Yifan and Ge, Chunjiang and Song, Jun and Song, Shiji and Zheng, Bo and Huang, Gao},
  booktitle={NeurIPS},
  year={2024},
}

@article{vvt,
  title={Vicinity vision transformer},
  author={Sun, Weixuan and Qin, Zhen and Deng, Hui and Wang, Jianyuan and Zhang, Yi and Zhang, Kaihao and Barnes, Nick and Birchfield, Stan and Kong, Lingpeng and Zhong, Yiran},
  journal={TPAMI},
  year={2023},
}

@inproceedings{gsva,
  title={Gsva: Generalized segmentation via multimodal large language models},
  author={Xia, Zhuofan and Han, Dongchen and Han, Yizeng and Pan, Xuran and Song, Shiji and Huang, Gao},
  booktitle={CVPR},
  year={2024}
}

@book{digital_image,
  title={Digital image processing algorithms and applications},
  author={Pitas, Ioannis},
  year={2000},
  publisher={John Wiley \& Sons}
}

@inproceedings{fno,
  title={Fourier Neural Operator for Parametric Partial Differential Equations},
  author={Li, Zongyi and Kovachki, Nikola Borislavov and Azizzadenesheli, Kamyar and Bhattacharya, Kaushik and Stuart, Andrew and Anandkumar, Anima and others},
  booktitle={ICLR},
  year={2021}
}

@inproceedings{gfn,
  title={Global filter networks for image classification},
  author={Rao, Yongming and Zhao, Wenliang and Zhu, Zheng and Lu, Jiwen and Zhou, Jie},
  booktitle={NeurIPS},
  year={2021}
}

@article{afno,
  title={Adaptive fourier neural operators: Efficient token mixers for transformers},
  author={Guibas, John and Mardani, Morteza and Li, Zongyi and Tao, Andrew and Anandkumar, Anima and Catanzaro, Bryan},
  journal={arXiv preprint arXiv:2111.13587},
  year={2021}
}

@inproceedings{aff,
  title={Adaptive frequency filters as efficient global token mixers},
  author={Huang, Zhipeng and Zhang, Zhizheng and Lan, Cuiling and Zha, Zheng-Jun and Lu, Yan and Guo, Baining},
  booktitle={ICCV},
  year={2023}
}

@inproceedings{freq_deblur,
  title={Efficient frequency domain-based transformers for high-quality image deblurring},
  author={Kong, Lingshun and Dong, Jiangxin and Ge, Jianjun and Li, Mingqiang and Pan, Jinshan},
  booktitle={CVPR},
  year={2023}
}

@inproceedings{fastervit,
  title={FasterViT: Fast Vision Transformers with Hierarchical Attention},
  author={Hatamizadeh, Ali and Heinrich, Greg and Yin, Hongxu and Tao, Andrew and Alvarez, Jose M and Kautz, Jan and Molchanov, Pavlo},
  booktitle={ICLR},
  year={2024},
}

@article{soft_plus,
  title={Softmax-free linear transformers},
  author={Lu, Jiachen and Zhang, Junge and Zhu, Xiatian and Feng, Jianfeng and Xiang, Tao and Zhang, Li},
  journal={IJCV},
  year={2024},
}

@inproceedings{vmamba,
  title={Vmamba: Visual state space model},
  author={Liu, Yue and Tian, Yunjie and Zhao, Yuzhong and Yu, Hongtian and Xie, Lingxi and Wang, Yaowei and Ye, Qixiang and Jiao, Jianbin and Liu, Yunfan},
  booktitle={NeurIPS},
  year={2024}
}

@inproceedings{localmamba,
  title={Localmamba: Visual state space model with windowed selective scan},
  author={Huang, Tao and Pei, Xiaohuan and You, Shan and Wang, Fei and Qian, Chen and Xu, Chang},
  booktitle={ECCVW},
  year={2024}
}

@inproceedings{cpe,
  title={Conditional Positional Encodings for Vision Transformers},
  author={Chu, Xiangxiang and Tian, Zhi and Zhang, Bo and Wang, Xinlong and Shen, Chunhua},
  booktitle={ICLR},
  year={2023}
}

@article{quadrangle,
  title={Vision transformer with quadrangle attention},
  author={Zhang, Qiming and Zhang, Jing and Xu, Yufei and Tao, Dacheng},
  journal={TPAMI},
  year={2024},
}

@inproceedings{slab,
  title={SLAB: Efficient Transformers with Simplified Linear Attention and Progressive Re-parameterized Batch Normalization},
  author={Guo, Jialong and Chen, Xinghao and Tang, Yehui and Wang, Yunhe},
  booktitle={ICML},
  year={2024},
}

@inproceedings{polaformer,
  title={PolaFormer: Polarity-aware Linear Attention for Vision Transformers},
  author={Meng, Weikang and Luo, Yadan and Li, Xin and Jiang, Dongmei and Zhang, Zheng},
  booktitle={ICLR},
  year={2025},
}

@inproceedings{inline,
  title={Bridging the divide: Reconsidering softmax and linear attention},
  author={Han, Dongchen and Pu, Yifan and Xia, Zhuofan and Han, Yizeng and Pan, Xuran and Li, Xiu and Lu, Jiwen and Song, Shiji and Huang, Gao},
  booktitle={NeurIPS},
  year={2024}
}

@inproceedings{agent_attention,
  title={Agent attention: On the integration of softmax and linear attention},
  author={Han, Dongchen and Ye, Tianzhu and Han, Yizeng and Xia, Zhuofan and Song, Shiji and Huang, Gao},
  booktitle={ECCV},
  year={2024},
}

@inproceedings{overlock,
  title={OverLoCK: An Overview-first-Look-Closely-next ConvNet with Context-Mixing Dynamic Kernels},
  author={Lou, Meng and Yu, Yizhou},
  booktitle={CVPR},
  year={2025}
}

@article{transxnet,
  title={TransXNet: learning both global and local dynamics with a dual dynamic token mixer for visual recognition},
  author={Lou, Meng and Zhang, Shu and Zhou, Hong-Yu and Yang, Sibei and Wu, Chuan and Yu, Yizhou},
  journal={TNNLS},
  year={2025},
}

@inproceedings{efficientvit,
  title={Efficientvit: Lightweight multi-scale attention for high-resolution dense prediction},
  author={Cai, Han and Li, Junyan and Hu, Muyan and Gan, Chuang and Han, Song},
  booktitle={ICCV},
  year={2023}
}

@phdthesis{wang2019kernel,
  title={Kernel learning for visual perception},
  author={Wang, Chen},
  year={2019},
  school={Columbia University}
}

@article{wang2025emulating,
  title={Emulating human-like adaptive vision for efficient and flexible machine visual perception},
  author={Wang, Yulin and Yue, Yang and Yue, Yang and Wang, Huanqian and Jiang, Haojun and Han, Yizeng and Ni, Zanlin and Pu, Yifan and Shi, Minglei and Lu, Rui and others},
  journal={Nature Machine Intelligence},
  year={2025},
}

\clearpage
\appendix

\section{BCCB Matrix}

Block Circulant matrix with Circulant Blocks (BCCB) is the 2D generalization of circulant matrix~\cite{davis1979circulant}. A BCCB matrix $B \in \mathbb{R}^{N \times N},N=H\times W$ has a block circulant structure with $H \times H$ blocks:
\begin{equation}
    B = 
    \begin{pmatrix}
    C_0 & C_1 & \cdots & C_{H-1} \\
    C_{H-1} & C_0 & \cdots & C_{H-2} \\
    \vdots & \vdots & \ddots & \vdots \\
    C_1 & C_2 & \cdots & C_0
    \end{pmatrix},
\end{equation}
where each block $C_i$ is a circulant matrix of size $W \times W$. Similar to the circulant matrix, the BCCB matrix $B$ is fully determined by its first row $b = [c_0, c_1, \dots, c_{HW-1}]$. 

Let $\hat{b},\hat{x}\in \mathbb{R}^{H \times W}$ be the 2D reshaped versions of $b,x$, respectively. We now prove that the multiplication $y=Bx$ is equivalent to the 2D circular cross-correlation between $\hat{b}$ and $\hat{x}$, i.e., the 2D global convolution with circular padding in deep learning. Notably, the concept of convolution in deep learning does not involve flipping the kernel, thus corresponding to the cross-correlation operation.

 We interpret the index \( 0 \leq k < N \) as a pair of two-dimensional coordinates \(( k_x, k_y)\), where \( k_x = \lfloor k / W \rfloor \) and \( k_y = k\bmod W  \). Using this mapping, the matrix entry $B_{i,j}$ can be written as $B_{(i_x,i_y),(j_x,j_y)}$ and $b_i$ can be written as $b_{(i_x,i_y)}$.

 Then, for the matrix entry $B_{i,j}=B_{(i_x,i_y),(j_x,j_y)}$, we can see that:
 \begin{itemize}
     \item The indices $i_x$ and $j_x$ (from $0$ to $H-1$) specify the \textbf{block-row} and \textbf{block-column}.
     \item The indices $i_y$ and $j_y$ (from $0$ to $W-1$) specify the \textbf{intra-block row} and \textbf{intra-block column}.
 \end{itemize}
 Thus, the element $B_{(i_x,i_y),(j_x,j_y)}$ is located in the row $i_y$ and column $j_y$ of the matrix block in the block row $i_x$ and block column $j_x$.

 Since the matrix $B$ has a block circulant structure, the block in block row $i_x$ and block column $j_x$ is given by $C_{(j_x-i_x) \bmod H}$.

 We can see that 
\begin{equation}
    \begin{split}
        B_{(i_x,i_y),(j_x,j_y)}&=(C_{(j_x-i_x)\bmod H})_{i_y,j_y}\\
        &=(C_{(j_x-i_x)\bmod H})_{(j_y-i_y)\bmod W}\\
        &=b_{((j_x-i_x)\bmod H,(j_y-i_y)\bmod W)}
    \end{split}
\end{equation}

We can express the matrix multiplication \( B x \), where \( B \in \mathbb{R}^{N \times N} \) and \( x \in \mathbb{R}^N \), as follows:
\begin{equation}
    \begin{split}
        (Bx)_{(i_x,i_y)} &= \sum_{j=0}^{N-1} B_{(i_x,i_y),(j_x,j_y)}  x_{(j_x,j_y)}\\
        &= \sum_{j_x=0}^{H-1} \sum_{j_y=0}^{W-1} B_{(i_x,i_y),(j_x,j_y)} x_{(j_x,j_y)}\\
        &= \sum_{j_x=0}^{H-1} \sum_{j_y=0}^{W-1} b_{((j_x-i_x)\bmod H,(j_y-i_y)\bmod W)} x_{(j_x,j_y)}
    \end{split}
\end{equation}


 Let $k_x = (j_x-i_x)\bmod H$ and $k_y = (j_y-i_y)\bmod W$. This implies $j_x = (i_x+k_x)\bmod H$ and $j_y = (i_y+k_y)\bmod W$. Substituting these into the summation yields:
 \begin{equation}
 {y}_{(i_x,i_y)} = \sum_{k_x=0}^{H-1} \sum_{k_y=0}^{W-1} b_{(k_x, k_y)}  x_{((i_x+k_x)\bmod H, (i_y+k_y)\bmod W)}
 \end{equation}
 This expression is the definition of the \textbf{2D circular cross-correlation} between the 2D kernel \({b}\) and the 2D signal \({x}\).

 We leverage the \textbf{2D Cross-Correlation Theorem}, which is a direct extension of the 1D theorem. It states that the 2D Fourier transform of a cross-correlation is the element-wise product of the first signal's conjugated 2D Fourier transform and the second signal's 2D Fourier transform:
 \begin{equation}
 \FFT_{\rm 2D}(Bx) = \conj{\FFT_{\rm 2D}({b})} \odot \FFT_{\rm 2D}({x})
 \end{equation}
 Thus,we have:
 \begin{equation}
 Bx = \IFFT_{\rm 2D}\left(\conj{\FFT_{\rm 2D}({b})} \odot \FFT_{\rm 2D}({x})\right)
 \end{equation}
 where $\FFT_{\rm 2D},\IFFT_{\rm 2D}$ denotes the 2D DFT and its inversion. For simplicity, the reshaping operations between 1D $\mathbb{R}^N$ sequences and 2D $\mathbb{R}^{H \times W}$ feature maps are not demonstrated, and we define the inputs/outputs of $\FFT_{\rm 2D},\IFFT_{\rm 2D}$ to be 1D sequences. Therefore, we can also compute it using the fast Fourier transform with a time complexity of $O(N \log N)$.

\begin{table*}[t!]
    \centering
    \scriptsize
    \setlength{\tabcolsep}{0.5mm}{
    \renewcommand\arraystretch{1.5}
    \begin{tabular}{c|c|c|c|c|c}
    \bottomrule
    \multicolumn{2}{c|}{CA-DeiT-T} & \multicolumn{2}{c|}{CA-DeiT-S} & \multicolumn{2}{c}{CA-DeiT-B}\\
    \cline{1-6}
    {CA Block} & DeiT Block& {CA Block} & DeiT Block& {CA Block} & DeiT Block \\
    \hline
    $\left[\!\!\! \begin{array}{c} {\rm \ res}  \ 14\!\times\! 14\ \ \\{\rm dim} \ 192 \\ {\rm head} \ 192\end{array} \!\!\! \right ] \!\!\times\! 12$ & None & $\left[\!\!\! \begin{array}{c} {\rm \ res}  \ 14\!\times\! 14\ \ \\{\rm dim} \ 384 \\ {\rm head} \ 384\end{array} \!\!\! \right ] \!\!\times\! 12$ & None & $\left[\!\!\! \begin{array}{c} {\rm \ res}  \ 14\!\times\! 14\ \ \\{\rm dim} \ 768 \\ {\rm head} \ 768\end{array} \!\!\! \right ] \!\!\times\! 12$ & None \\
    \toprule
    \end{tabular}}
    \caption{Architectures of CA-DeiT models.}
    \label{tab:model_deit}
\end{table*}

\begin{table*}[t!]
    \centering
    \scriptsize
    \setlength{\tabcolsep}{3.5mm}{
    \renewcommand\arraystretch{1.5}
    \begin{tabular}{c|c|c|c|c|c}
    \bottomrule
    \multirow{2}*{Stage} & \multirow{2}*{Output} & \multicolumn{2}{c|}{CA-PVT-T} & \multicolumn{2}{c}{CA-PVT-S}\\
    \cline{3-6}
    & & {CA Block} & PVT Block & {CA Block} & PVT Block \\
    \hline
    \multirow{4}*{res1} & \multirow{4}*{$56\times 56$} & \multicolumn{4}{c}{Conv4×4, stride=4, 64, LN}\\
    \cline{3-6}
    && $\left[\!\!\! \begin{array}{c} {\rm \ win}  \ 56\!\times\! 56\ \ \\{\rm dim} \ 64 \\ {\rm head} \ 64  \end{array} \!\!\! \right ] \!\!\times\! 2$ & None & $\left[\!\!\! \begin{array}{c} {\rm \ win}  \ 56\!\times\! 56\ \ \\{\rm dim} \ 64 \\ {\rm head} \ 64  \end{array} \!\!\! \right ] \!\!\times\! 3$ & None \\
    \hline
    \multirow{4}*{res2} & \multirow{4}*{$28\times 28$} & \multicolumn{4}{c}{Conv2×2, stride=2, 128, LN}\\
    \cline{3-6}
    && $\left[\!\!\! \begin{array}{c} {\rm \ win}  \ 28\!\times\! 28\ \ \\{\rm dim} \ 128 \\ {\rm head} \ 128 \end{array} \!\!\! \right ] \!\!\times\! 2$ & None & $\left[\!\!\! \begin{array}{c} {\rm \ win}  \ 28\!\times\! 28\ \ \\{\rm dim} \ 128 \\ {\rm head} \ 128 \end{array} \!\!\! \right ] \!\!\times\! 4$ & None \\
    \hline
    \multirow{4}*{res3} & \multirow{4}*{$14\times 14$} & \multicolumn{4}{c}{Conv2×2, stride=2, 320, LN}\\
    \cline{3-6}
    & & None& $\left[\!\!\! \begin{array}{c} {\rm \ win}  \ 14\!\times\! 14\ \ \\{\rm dim} \ 320 \\ {\rm head} \ 5 \end{array} \!\!\! \right ] \!\!\times\! 2$ & None & $\left[\!\!\! \begin{array}{c} {\rm \ win}  \ 14\!\times\! 14\ \ \\{\rm dim} \ 320 \\ {\rm head} \ 5 \end{array} \!\!\! \right ] \!\!\times\! 6$  \\
    \hline
    \multirow{4}*{res4} & \multirow{4}*{$7\times 7$} & \multicolumn{4}{c}{Conv2×2, stride=2, 512, LN}\\
    \cline{3-6}
    & & None & $\left[\!\!\! \begin{array}{c} {\rm \ win}  \ 7\!\times\! 7\ \ \\{\rm dim} \ 512 \\ {\rm head} \ 8 \end{array} \!\!\! \right ] \!\!\times\! 2$ & None & $\left[\!\!\! \begin{array}{c} {\rm \ win}  \ 7\!\times\! 7\ \ \\{\rm dim} \ 512 \\ {\rm head} \ 8 \end{array} \!\!\! \right ] \!\!\times\! 3$ \\
    \toprule
    \end{tabular}}
    \caption{Architectures of CA-PVT models (Part1).}
    \label{tab:model_pvt-1}
\end{table*}

\begin{table*}[t!]
    \centering
    \scriptsize
    \setlength{\tabcolsep}{3.5mm}{
    \renewcommand\arraystretch{1.5}
    \begin{tabular}{c|c|c|c|c|c}
    \bottomrule
    \multirow{2}*{Stage} & \multirow{2}*{Output} & \multicolumn{2}{c|}{CA-PVT-M} & \multicolumn{2}{c|}{CA-PVT-L}\\
    \cline{3-6}
    & & {CA Block} & PVT Block & {CA Block} & PVT Block \\
    \hline
    \multirow{4}*{res1} & \multirow{4}*{$56\times 56$} & \multicolumn{4}{c}{Conv4×4, stride=4, 64, LN}\\
    \cline{3-6}
    && $\left[\!\!\! \begin{array}{c} {\rm \ win}  \ 56\!\times\! 56 \ \ \\{\rm dim} \ 64 \\ {\rm head} \ 64  \end{array} \!\!\! \right ] \!\!\times\! 3$ & None & $\left[\!\!\! \begin{array}{c} {\rm \ win}  \ 56\!\times\! 56\ \ \\{\rm dim} \ 64 \\ {\rm head} \ 64  \end{array} \!\!\! \right ] \!\!\times\! 3$ & None \\
    \hline
    \multirow{4}*{res2} & \multirow{4}*{$28\times 28$} & \multicolumn{4}{c}{Conv2×2, stride=2, 128, LN}\\
    \cline{3-6}
    && $\left[\!\!\! \begin{array}{c} {\rm \ win}  \ 28\!\times\! 28\ \ \\{\rm dim} \ 128 \\ {\rm head} \ 128 \end{array} \!\!\! \right ] \!\!\times\! 4$ & None & $\left[\!\!\! \begin{array}{c} {\rm \ win}  \ 28\!\times\! 28\ \ \\{\rm dim} \ 128 \\ {\rm head} \ 128 \end{array} \!\!\! \right ] \!\!\times\! 8$ & None \\
    \hline
    \multirow{4}*{res3} & \multirow{4}*{$14\times 14$} & \multicolumn{4}{c}{Conv2×2, stride=2, 320, LN}\\
    \cline{3-6}
    & & None& $\left[\!\!\! \begin{array}{c} {\rm \ win}  \ 14\!\times\! 14\ \ \\{\rm dim} \ 320 \\ {\rm head} \ 5 \end{array} \!\!\! \right ] \!\!\times\! 18$ & None & $\left[\!\!\! \begin{array}{c} {\rm \ win}  \ 14\!\times\! 14\ \ \\{\rm dim} \ 320 \\ {\rm head} \ 5 \end{array} \!\!\! \right ] \!\!\times\! 27$  \\
    \hline
    \multirow{4}*{res4} & \multirow{4}*{$7\times 7$} & \multicolumn{4}{c}{Conv2×2, stride=2, 512, LN}\\
    \cline{3-6}
    & & None&$\left[\!\!\! \begin{array}{c} {\rm \ win}  \ 7\!\times\! 7\ \ \\{\rm dim} \ 512 \\ {\rm head} \ 8 \end{array} \!\!\! \right ] \!\!\times\! 3$ & None & $\left[\!\!\! \begin{array}{c} {\rm \ win}  \ 7\!\times\! 7\ \ \\{\rm dim} \ 512 \\ {\rm head} \ 8 \end{array} \!\!\! \right ] \!\!\times\! 3$  \\
    \toprule
    \end{tabular}}
    \caption{Architectures of CA-PVT models (Part2).}
    \label{tab:model_pvt-2}
\end{table*}

\begin{table*}[t!]
    \centering
    \scriptsize
    \setlength{\tabcolsep}{1.5mm}{
    \renewcommand\arraystretch{1.5}
    \begin{tabular}{c|c|c|c|c|c|c|c}
    \bottomrule
    \multirow{2}*{Stage} & \multirow{2}*{Output} & \multicolumn{2}{c|}{CA-Swin-T} & \multicolumn{2}{c|}{CA-Swin-S} & \multicolumn{2}{c}{CA-Swin-B}\\
    \cline{3-8}
    & & {CA Block} & Swin Block & {CA Block} & Swin Block& {CA Block} & Swin Block\\
    \hline
    \multirow{4}*{res1} & \multirow{4}*{$56\times 56$} & \multicolumn{2}{c|}{concat $4\times 4$, 96, LN} & \multicolumn{2}{c|}{concat $4\times 4$, 96, LN} & \multicolumn{2}{c}{concat $4\times 4$, 128, LN}\\
    \cline{3-8}
    && $\left[\!\!\! \begin{array}{c} {\rm \ win}  \ 56\!\times\! 56\ \ \\{\rm dim} \ 96 \\ {\rm head} \ 96  \end{array} \!\!\! \right ] \!\!\times\! 2$ & None & $\left[\!\!\! \begin{array}{c} {\rm \ win}  \ 56\!\times\! 56\ \ \\{\rm dim} \ 96 \\ {\rm head} \ 96  \end{array} \!\!\! \right ] \!\!\times\! 2$ & None & $\left[\!\!\! \begin{array}{c} {\rm \ win}  \ 56\!\times\! 56\ \ \\{\rm dim} \ 128 \\ {\rm head} \ 128  \end{array} \!\!\! \right ] \!\!\times\! 2$ & None\\
    \hline
    \multirow{4}*{res2} & \multirow{4}*{$28\times 28$} & \multicolumn{2}{c|}{concat $2\times 2$, 192, LN} & \multicolumn{2}{c|}{concat $2\times 2$, 192, LN} & \multicolumn{2}{c}{concat $2\times 2$, 256, LN}\\
    \cline{3-8}
    && $\left[\!\!\! \begin{array}{c} {\rm \ win}  \ 28\!\times\! 28\ \ \\{\rm dim} \ 192 \\ {\rm head} \ 192 \end{array} \!\!\! \right ] \!\!\times\! 2$ & None & $\left[\!\!\! \begin{array}{c} {\rm \ win}  \ 28\!\times\! 28\ \ \\{\rm dim} \ 192 \\ {\rm head} \ 192 \end{array} \!\!\! \right ] \!\!\times\! 2$ & None & $\left[\!\!\! \begin{array}{c} {\rm \ win}  \ 28\!\times\! 28\ \ \\{\rm dim} \ 256 \\ {\rm head} \ 256 \end{array} \!\!\! \right ] \!\!\times\! 2$ & None\\
    \hline
    \multirow{4}*{res3} & \multirow{4}*{$14\times 14$} & \multicolumn{2}{c|}{concat $2\times 2$, 384, LN} & \multicolumn{2}{c|}{concat $2\times 2$, 384, LN} & \multicolumn{2}{c}{concat $2\times 2$, 512, LN}\\
    \cline{3-8}
    && None & $\left[\!\!\! \begin{array}{c} {\rm \ win}  \ 7\!\times\! 7\ \ \\{\rm dim} \ 384 \\ {\rm head} \ 12\end{array} \!\!\! \right ] \!\!\times\! 6$ & None & $\left[\!\!\! \begin{array}{c} {\rm \ win}  \ 7\!\times\! 7\ \ \\{\rm dim} \ 384 \\ {\rm head} \ 12\end{array} \!\!\! \right ] \!\!\times\! 18$ &None & $\left[\!\!\! \begin{array}{c} {\rm \ win}  \ 7\!\times\! 7\ \ \\{\rm dim} \ 512 \\ {\rm head} \ 16\end{array} \!\!\! \right ] \!\!\times\! 18$\\
    \hline
    \multirow{4}*{res4} & \multirow{4}*{$7\times 7$} & \multicolumn{2}{c|}{concat $2\times 2$, 768, LN} & \multicolumn{2}{c|}{concat $2\times 2$, 768, LN} & \multicolumn{2}{c}{concat $2\times 2$, 1024, LN}\\
    \cline{3-8}
    & & None& $\left[\!\!\! \begin{array}{c} {\rm \ win}  \ 7\!\times\! 7\ \ \\{\rm dim} \ 768 \\ {\rm head} \ 24\end{array} \!\!\! \right ] \!\!\times\! 2$ & None & $\left[\!\!\! \begin{array}{c} {\rm \ win}  \ 7\!\times\! 7\ \ \\{\rm dim} \ 768 \\ {\rm head} \ 24\end{array} \!\!\! \right ] \!\!\times\! 2$ & None & $\left[\!\!\! \begin{array}{c} {\rm \ win}  \ 7\!\times\! 7\ \ \\{\rm dim} \ 1024 \\ {\rm head} \ 32\end{array} \!\!\! \right ] \!\!\times\! 2$\\
    \toprule
    \end{tabular}}
    \caption{Architectures of CA-Swin models.}
    \label{tab:model_swin}
\end{table*}

\begin{table*}[t!]
    \centering
    \scriptsize
    \setlength{\tabcolsep}{1.5mm}{
    \renewcommand\arraystretch{1.5}
    \begin{tabular}{c|c|c|c|c|c|c|c} 
    \toprule
    \multirow{2}*{Stage} & \multirow{2}*{Output} & \multicolumn{2}{c|}{CAT-T} & \multicolumn{2}{c|}{CAT-S} & \multicolumn{2}{c}{CAT-B}\\
    \cline{3-8}
    & & {CA Block} & Attn Block & {CA Block} & Attn Block & {CA Block} & Attn Block\\
    \hline
    \multirow{4}*{res1} & \multirow{4}*{$56\times 56$} & \multicolumn{2}{c|}{Stem, stride 4, 64, BN} & \multicolumn{2}{c|}{Stem, stride 4, 64, BN} & \multicolumn{2}{c}{Stem, stride 4, 96, BN}\\
    \cline{3-8}
    && $\left[\!\!\begin{array}{c} {\rm win} \ 56\!\times\!56 \\ {\rm dim} \ 64 \\{\rm head}\ 64 \end{array}\!\!\right]\!\!\times\!1$ & None & $\left[\!\!\begin{array}{c} {\rm win} \ 56\!\times\!56 \\ {\rm dim} \ 64\\{\rm head}\ 64 \end{array}\!\!\right]\!\!\times\!2$ & None & $\left[\!\!\begin{array}{c} {\rm win} \ 56\!\times\!56 \\ {\rm dim} \ 96 \\{\rm head}\ 96\end{array}\!\!\right]\!\!\times\!2$ & None \\
    \hline
    \multirow{4}*{res2} & \multirow{4}*{$28\times 28$} & \multicolumn{2}{c|}{Conv 3$\times$3, stride 2, 128, BN} & \multicolumn{2}{c|}{Conv 3$\times$3, stride 2, 128, BN} & \multicolumn{2}{c}{Conv 3$\times$3, stride 2, 192, BN}\\
    \cline{3-8}
    && $\left[\!\!\begin{array}{c} {\rm win} \ 28\!\times\!28 \\ {\rm dim} \ 128\\{\rm head}\ 128 \end{array}\!\!\right]\!\!\times\!2$ & None & $\left[\!\!\begin{array}{c} {\rm win} \ 28\!\times\!28 \\ {\rm dim} \ 128\\{\rm head}\ 128 \end{array}\!\!\right]\!\!\times\!4$ & None & $\left[\!\!\begin{array}{c} {\rm win} \ 28\!\times\!28 \\ {\rm dim} \ 192\\{\rm head}\ 192 \end{array}\!\!\right]\!\!\times\!4$ & None \\
    \hline
    \multirow{4}*{res3} & \multirow{4}*{$14\times 14$} & \multicolumn{2}{c|}{Conv 3$\times$3, stride 2, 320, BN} & \multicolumn{2}{c|}{Conv 3$\times$3, stride 2, 320, BN} & \multicolumn{2}{c}{Conv 3$\times$3, stride 2, 448, BN}\\
    \cline{3-8}
    && None & $\left[\!\!\begin{array}{c} {\rm win} \ 14\!\times\!14 \\ {\rm dim} \ 320 \\ {\rm head} \ 10 \end{array}\!\!\right]\!\!\times\!9$ & None & $\left[\!\!\begin{array}{c} {\rm win} \ 14\!\times\!14 \\ {\rm dim} \ 320 \\ {\rm head} \ 10 \end{array}\!\!\right]\!\!\times\!18$ & None & $\left[\!\!\begin{array}{c} {\rm win} \ 14\!\times\!14 \\ {\rm dim} \ 448 \\ {\rm head} \ 14 \end{array}\!\!\right]\!\!\times\!18$ \\
    \hline
    \multirow{4}*{res4} & \multirow{4}*{$7\times 7$} & \multicolumn{2}{c|}{Conv 3$\times$3, stride 2, 512, BN} & \multicolumn{2}{c|}{Conv 3$\times$3, stride 2, 512, BN} & \multicolumn{2}{c}{Conv 3$\times$3, stride 2, 640, BN}\\
    \cline{3-8}
    && None & $\left[\!\!\begin{array}{c} {\rm win} \ 7\!\times\!7 \\ {\rm dim} \ 512 \\ {\rm head} \ 16 \end{array}\!\!\right]\!\!\times\!4$ & None & $\left[\!\!\begin{array}{c} {\rm win} \ 7\!\times\!7 \\ {\rm dim} \ 512 \\ {\rm head} \ 16 \end{array}\!\!\right]\!\!\times\!8$ & None & $\left[\!\!\begin{array}{c} {\rm win} \ 7\!\times\!7 \\ {\rm dim} \ 640 \\ {\rm head} \ 20 \end{array}\!\!\right]\!\!\times\!8$ \\
    \bottomrule
    \end{tabular}}
    \caption{Architectures of CAT models.}
    \label{tab:model_CAT}
\end{table*}

\section{Model Architectures}

We provide the architectures of CA-DeiT, CA-PVT, CA-Swin and CAT in \cref{tab:model_deit}, \cref{tab:model_pvt-1}, \cref{tab:model_pvt-2}, \cref{tab:model_swin} and \cref{tab:model_CAT}. Our circulant attention serves as a plug-in module. We simply replace the original attention module with the proposed circulant attention, while keeping the other network components unchanged. To introduce positional information, we employ conditional positional encodings~\cite{cpe}. Since our method benefits from $N\log N$ complexity, it is possible to directly process a high-resolution feature map in the early stages without incurring high computational cost. Therefore, for hierarchical models, we mainly restrict the attention replacement to the first two stages.

\end{document}